\documentclass[runningheads]{tsd1108a}

\usepackage{graphicx}
\usepackage{todonotes}
\usepackage{adjustbox}
\usepackage{booktabs}
\usepackage{arydshln}
\usepackage{pgfplots}
\usepackage{subcaption}
\usepackage{amsmath, amssymb}
\usepackage{multirow}
\usepackage{xcolor}
\usepackage[para]{footmisc}

\pgfplotsset{compat=1.17}

\makeatletter
\def\adl@drawiv#1#2#3{%
        \hskip.5\tabcolsep
        \xleaders#3{#2.5\@tempdimb #1{1}#2.5\@tempdimb}%
                #2\z@ plus1fil minus1fil\relax
        \hskip.5\tabcolsep}
\newcommand{\cdashlinelr}[1]{%
  \noalign{\vskip\aboverulesep
           \global\let\@dashdrawstore\adl@draw
           \global\let\adl@draw\adl@drawiv}
  \cdashline{#1}
  \noalign{\global\let\adl@draw\@dashdrawstore
           \vskip\belowrulesep}}
\makeatother

\begin{document}

\title{Linear Transformations for Cross-lingual Sentiment Analysis
}
%
%

\author{Pavel P\v{r}ib\'{a}\v{n}\inst{1,2}\orcidID{0000-0002-8744-8726} \and Jakub \v{S}m\'{i}d\inst{1}\orcidID{0000-0002-4492-5481} \and Adam Mi\v{s}tera \inst{1,2} \and Pavel Kr\'{a}l\inst{1}\orcidID{0000-0002-3096-675X}} 

 \institute{University of West Bohemia\\
          Faculty of Applied Sciences, Department of Computer Science and Engineering\\
          \and
          NTIS -- New Technologies for the Information Society\\
          Univerzitni 8, 301 00 Plze\v{n}, Czech Republic\\
          \email{\{pribanp,amistera,pkral\}@kiv.zcu.cz, biba10@students.zcu.cz}
            \tt {\url{https://nlp.kiv.zcu.cz}} \\
          }

\maketitle              

\renewcommand{\vec}[1]{\ensuremath{\mathbf{#1}}}

\begin{abstract}
This paper deals with cross-lingual sentiment analysis in Czech, English and French languages. We perform zero-shot cross-lingual classification using five linear transformations combined with LSTM and CNN based classifiers. We compare the performance of the individual transformations, and in addition, we confront the transformation-based approach with existing state-of-the-art BERT-like models. 
We show that the pre-trained embeddings from the target domain are crucial to improving the cross-lingual classification results, unlike in the monolingual classification, where the effect is not so distinctive.

\keywords{sentiment analysis  \and cross-lingual \and linear transformation \and neural networks \and semantic space transformation \and classification}
\end{abstract}

\section{Introduction}



\par Sentiment analysis
(SA) is an essential task in the natural language processing (NLP) field and a lot of research interest has been devoted to this topic, especially in monolingual settings for English. However, cross-lingual sentiment analysis (CLSA) has been much less studied. Most of the approaches to SA require annotated data. CLSA aims to enable knowledge transfer between languages with enough data and languages with less or without annotated data (low-resource languages), thus allowing to run SA in these languages.

\par We can divide the existing approaches for CLSA into three groups. Machine translation can be used to translate annotated data into the target language and then the model is trained using the translated data. Secondly, the multilingual versions of pre-trained BERT-like models such as mBERT \cite{devlin-etal-2019-bert} or XLM-R \cite{xlm-r}
are applicable to CLSA. These models can be fine-tuned with annotated data from high-resource languages, usually English. Classification in the target language, such as Czech, is then performed without any training data \cite{priban-steinberger-2021-multilingual}. The third approach uses linear transformations and other methods to transfer knowledge between languages.
Usually, the linear transformations align semantic spaces \cite{ruder2019survey} in different languages into one common space. The common space (word embeddings) is then used during the training of a neural network, namely Long Short-Term Memory (LSTM) or Convolutional Neural Network (CNN).

\par As shown in \cite{priban-steinberger-2021-multilingual},
multilingual BERT-like models achieve SotA results.
Their drawback is that they typically require much more resources and computational power in terms of memory and GPU performance than previous approaches. These greater requirements cause that they usually have to be fine-tuned on expensive and specialized GPUs with high electricity consumption. On the other hand, the approaches based on linear transformations and neural networks (e.g., CNN) can be easily trained on a standard computer using only a CPU. As we show in our paper, these cheaper cross-lingual approaches achieve competitive results in comparison to the multilingual BERT-like models. 

\par In this paper, we focus on the \textit{Sentiment Classification}\footnote{Here, we consider sentiment analysis and sentiment classification as the same task.} task (also known as \textit{Polarity detection}) in cross-lingual settings for Czech, English and French. We perform zero-shot cross-lingual sentiment classification in the three languages. We compare the performance of five linear transformations for aligning semantic spaces in combination with two neural network models: CNN and LSTM. We show that the source of embeddings is a very important aspect when using linear transformations for the CLSA. Our experiments reveal that pre-trained in-domain embeddings can significantly improve (even more than 10\%) cross-lingual classification results, unlike in monolingual classification, where the difference is only about 1\%--2\%. We compare our results with the available cross-lingual SotA results based on multilingual BERT-like models.
In addition, to the best of our knowledge, none of the previous works applied linear transformations to the task of CLSA for the Czech language. We release all our resources and source codes\footnote{\url{https://github.com/pauli31/linear-transformation-4-cs-sa}\label{fnlabel}}.

\par Our main contributions are the following: 1) We compare the performance of five linear transformations on the task of CLSA and compare them with the available cross-lingual SotA results. 2) We show that the source of data for the embeddings used in the linear transformations is crucial for the CLSA task.




\section{Related Work}
Cross-lingual sentiment analysis has been moderately studied in recent years, but less attention has been paid to this research subtopic in comparison to the monolingual task.
The approaches proposed in~\cite{balahur-turchi-2012-multilingual,zhou-etal-2016-attention-based,rnn-based-ethem-2018}, the recent state-of-the-art BERT-like models applications \cite{barriere-balahur-2020-improving,priban-steinberger-2021-multilingual,zhang-etal-2021-cross,thakkar2021multi}, cross-lingual word embeddings and linear transformation
can be applied to tackle this task. Most of the following cited works do not compare the linear transformations and their effect on CLSA performance in such detail as our work does.

\par Authors in \cite{jain-batra-2015-cross} used a recursive autoencoder architecture and sentence aligned corpora of English and Hindi for CLSA and evaluated the system on the Hindi movie reviews dataset. A method specifically for sentiment cross-lingual word embeddings was proposed in \cite{zhou-etal-2015-learning-bilingual}. The authors trained an SVM classifier based on embeddings for the polarity classification task. In \cite{barnes-etal-2016-exploring},  multiple techniques for cross-lingual aspect-based sentiment classification were compared, including the one from \cite{mikolov2013exploiting}. Also, in \cite{abdalla-hirst-2017-cross}, the authors experimented with the linear transformation method from \cite{mikolov2013exploiting} on English, Spanish and Chinese. 
In \cite{barnes-etal-2018-bilingual}, an approach for training bilingual sentiment word embeddings is presented. The embeddings are jointly optimized to represent (a) semantic information in the source and target languages using a small bilingual dictionary and (b) sentiment information obtained from the source language only.
A cross-lingual algorithm using CNN is shown in \cite{dong2018cross} and evaluated on nine languages. Authors of \cite{chen-etal-2018-adversarial} trained an adversarial neural network with bilingual embeddings for polarity classification in Chinese and Arabic using only English train data.
In both  \cite{persian-sentiment-2020,kuriyozov-etal-2020-cross}, the authors used linear transformations for Persian and Turkish.

\section{Experimental Setup}
In this section, we describe the data, linear transformations and the models that we used for the experiments. We also cover the process of building bilingual dictionaries needed for the linear transformations along with the methodology of our zero-shot cross-lingual experiments.

\subsection{Data}
We use four publicly available datasets with binary polarity labels, i.e., \textit{positive} and \textit{negative} from the movie reviews domain. For Czech, we use the \textbf{CSFD} dataset of movie reviews introduced in \cite{habernal-etal-2013-sentiment}. It is built from 90k reviews from the Czech movie database\footnote{\url{https://www.csfd.cz}} that were downloaded and annotated according to their star rating (0–1 stars as \textit{negative}, 2-3 stars as \textit{neutral}, 4–5 stars as \textit{positive}). However, we use only the examples labeled as \textit{positive} or \textit{negative}. We use the data split from \cite{priban-steinberger-2021-multilingual}. The French \textbf{Allocine} \cite{allocine} dataset consists of 100k positive and 100k negative reviews.
The dataset was scraped from the Allociné\footnote{\url{https://www.allocine.fr}} website and annotated the same way as the CSFD dataset. English \textbf{IMDB} \cite{maas-etal-2011-learning-imdb} dataset includes 50k movie reviews obtained from the Internet Movie Database\footnote{\url{https://www.imdb.com}} with \textit{positive} and \textit{negative} classes.
We randomly selected 2.5k examples from the training part as development data. The second English \textbf{SST-2} \cite{socher-etal-2013-recursive-sst} dataset contains almost 12k manually annotated movie reviews into two categories, see Table \ref{tab:datasets}.

\begin{table}[ht!]
\caption{Dataset statistics.} \label{tab:datasets}
\setlength{\tabcolsep}{3pt}
\begin{adjustbox}{width=0.85\linewidth,center}

\begin{tabular}{lrrrlrrrlrrrlrrr} \toprule
        & \multicolumn{3}{c}{\textbf{CSFD} (Czech)}                                              &                      & \multicolumn{3}{c}{\textbf{IMDB} (English)}                                              &                      & \multicolumn{3}{c}{\textbf{SST-2} (English)}                                             &                      & \multicolumn{3}{c}{\textbf{Allocine} (French)}                                          \\ \cline{2-4} \cline{6-8} \cline{10-12} \cline{14-16} 
         & \multicolumn{1}{c}{train} & \multicolumn{1}{c}{dev} & \multicolumn{1}{c}{test} & \multicolumn{1}{c}{} & \multicolumn{1}{c}{train} & \multicolumn{1}{c}{dev} & \multicolumn{1}{c}{test} & \multicolumn{1}{c}{} & \multicolumn{1}{c}{train} & \multicolumn{1}{c}{dev} & \multicolumn{1}{c}{test} & \multicolumn{1}{c}{} & \multicolumn{1}{c}{train} & \multicolumn{1}{c}{dev} & \multicolumn{1}{c}{test} \\ \midrule 
Pos. & 22,117                    & 2,456                   & 6,324                    &                      & 11,242                    & 1,258                   & 12,500                   &                      & 3,610                     & 444                   & 909                    &                      & 79,413                    & 9,796                   & 9,592                    \\
Neg. & 21,441                    & 2,399                   & 5,876                    &                      & 11,258                    & 1,242                   & 12,500                   &                      & 3,310                     & 428                   & 912                    &                      & 80,587                    & 10,204                  & 10,408                   \\ \hdashline
Tot.    & 43,558                    & 4,855                   & 12,200                   &                      & 22,500                    & 2,500                   & 25,000                   &                      & 6,920                     & 872                   & 1,821                    &                      & 160,000                   & 20,000                  & 20,000                 \\ \bottomrule
\end{tabular}

\end{adjustbox}
\end{table}

\par The linear transformations require bilingual dictionaries to align the semantic spaces, see Section \ref{sec:cross-sentiment}. With Google Translate,
we translated the 40k most common words from the CSFD dataset into English and French to obtain the required dictionaries. We repeat the process for IMDB and Allocine datasets.
We also manually fixed some translation errors made by the translator.


\subsection{Linear Transformations}
\label{sec:linear_transformations}
We use linear transformations to create a bilingual semantic space. We align one semantic space (word embeddings) into the second semantic space in a different language. In such bilingual word embeddings, semantically similar words have similar vector representations across the two languages. Thanks to this property, we can use the bilingual space to train a neural network with a cross-lingual ability. The goal of the linear transformation is to find a transformation matrix $\vec{W}^{s \rightarrow  t} \in \mathbb{R}^{d \times d}$ that transforms vector space $\vec{X}^{s}$ of the source language into the vector space $\vec{X}^{t}$ of the target language by the following matrix multiplication:
\begin{equation}
  \vec{\hat{X}}^s =   \vec{W}^{s \rightarrow t} \vec{X}^s,
\end{equation}
where $\vec{\widehat{X}}^{s}$ is the transformed source vector space in the target space. Both matrices $\vec{X}^{t}$ and $\vec{X}^{s}$ contain $n$ vectors that correspond to translated pairs of words (called \textit{seed words}) in dictionary $D$. Any word vector that is not present in the source matrix $\vec{X}^{s}$ but comes from the same space can be transformed into the target space by multiplication of the transformation matrix $\vec{W}^{s \rightarrow  t}$.

\par We selected five transformation methods, the first was proposed in \cite{mikolov2013exploiting}, where the transformation matrix is estimated by minimizing the mean squared error (\textbf{MSE}) between the pairs of vectors $(\vec{x}_i^{s}, \vec{x}_i^{t})$ for the corresponding words from the dictionary $D$, as follows:
\begin{equation}
    MSE = \sum_{i=1}^{n}\left \|  \vec{W}^{s \rightarrow t}\vec{x}_{i}^{s} - \vec{x}_{i}^{t}\right \|^2.
\end{equation}

\par The second transformation is called Orthogonal (\textbf{Orto}) constraints the transformation matrix $\vec{W}^{s \rightarrow  t}$ to be \textit{orthogonal}\footnote{Matrix $\vec{W}$ is orthogonal when it is a square matrix and the columns and rows are orthonormal vectors ($\vec{W}^{\mathsf{T}}\vec{W} = \vec{W}\vec{W}^{\mathsf{T}} = I$, where $I$ is the identity matrix).}. The orthogonal transformation has the same objective as MSE.
The optimal transformation matrix $\vec{W}^{s \rightarrow t}$ can be computed as follows:
\begin{equation}
    \vec{W}^{s \rightarrow t} = \vec{V}\vec{U}^\mathsf{T},
\end{equation}
where matrices $\vec{V}$ and $\vec{U}$ are computed by \textit{Singular Value Decomposition} (SVD) of $\vec{X}^{t^\mathsf{T}}\vec{X}^s = \vec{U}\vec{\Sigma} \vec{V}^{\mathsf{T}}$ as shown in \cite{artetxe-etal-2016-learning}. The orthogonality constraint causes the transformation does not squeeze or re-scale the transformed space. It only rotates the space, thus it preserves most of the relationships of its elements (in our case, it is important that orthogonal transformation preserves angles between the words, so it preserves similarity between words in the transformed space).

\par The third method is based on \textit{Canonical Correlation Analysis} (\textbf{CCA}). The method aligns both monolingual vector spaces $\vec{X}^{s}$ and $\vec{X}^{t}$ to a third shared space represented by matrix $\vec{Y}^o$ \cite{ruder2019survey}. CCA computes two transformation matrices $\vec{W}^{s \rightarrow o}$ for the source language and $\vec{W}^{t \rightarrow o}$ for the target language to map the spaces into one shared space $\vec{Y}^o$. The transformation matrices can be computed analytically \cite{hardoon2004canonical} using SVD. Using the approach from \cite{ammar2016massively}, the transformation matrix $ \vec{W}^{s \rightarrow t} $ can be computed as follows:

\begin{equation}
     \vec{W}^{s \rightarrow t} = \vec{W}^{s \rightarrow o} (\vec{W}^{t \rightarrow o})^{-1}.
\end{equation}







\par The \textit{Ranking Transformation} (\textbf{Rank}) \cite{lazaridou-etal-2015-hubness} uses \textit{max-margin hinge loss} (MML) instead of MSE to reduce the \textit{hubness} problem \cite{ruder2019survey}. The idea of this method is to rank correct translations of word $w_i$ (i.e., vectors $\vec{x}_i^{s}$ and $\vec{x}_i^{t}$) higher than random translation (negative example) of word $w_i$ (i.e., vectors $\vec{x}_i^{s}$ and $\vec{x}_j^{t}$).

\par The last \textit{Orthogonal Ranking Transformation} (\textbf{Or-Ra}) \cite{BRYCHCIN2020104819} combines orthogonal and ranking transformations. The method tries to keep the transformation matrix $\vec{W}^{s \rightarrow t}$ orthogonal and reduce hubness at once, see \cite{BRYCHCIN2020104819} for the objective function and details.
Based on our experiments
and our empirical experiences, we decided to set dictionary size to 20k word pairs in every experiment.

\subsection{Neural Network Models}
For our experiments, we implement
a CNN inspired by \cite{kim2014convolutional} with one convolutional layer on top of pre-trained word embeddings. The first layer (embeddings layer) maps the input sentence of length \textit{n} to an $n\times d$ dimensional matrix, where $d = 300$ is the dimension of the word embeddings. We then use 1-dimensional convolution with filter sizes of 2, 3 and 4 (256 filters of each size) to extract features from the sentence. Next, we apply the ReLU activation and use max-over-time-pooling.
After the pooling, we concatenate these scalars into a vector, which is then fed to a fully-connected layer to compute the prediction scores for the classes. The class with the highest score is selected as the prediction. As a regularization, we use a dropout \cite{dropout} of 0.5 before the fully-connected layer.

\par We also train neural network based on the Bidirectional LSTM (BiLSTM).
Our model consists of an embedding layer that again maps the input words to 300-dimensional input vectors. These vectors then pass to two BiLSTM layers, each with $512$ units (hidden size). The output of the last BiLSTM layer is fed into the final fully-connected layer that computes prediction scores. We use a dropout of 0.5 before the fully-connected layer.

\par For our experiments, we use two types of pre-trained fastText \cite{bojanowski2016enriching} embeddings: (a) the existing\footnote{Available from \url{https://fasttext.cc/docs/en/crawl-vectors.html}} fastText embeddings. 
(b) In-domain embeddings trained by us on the text from the sentiment datasets. 
For English, we concatenate the texts from the SST-2 and IMDB datasets. We train embeddings for words with a minimum frequency of 5. 
During the training of the neural network models, we freeze the embeddings layer, meaning that we do not allow the embeddings to be fine-tuned during training.
For out-of-vocabulary words, we utilize the ability of fastText embeddings to generate a word vector. We train our model using Adam \cite{kingma2014adam} with constant learning of 1e-3 or 1e-4. For some experiments, we use linear learning rate decay.
We use a batch size of 32 and we train all our models for at most 10 epochs. We do not restrict the maximum input sequence length.



\subsection{Cross-lingual Sentiment Classification}
\label{sec:cross-sentiment}
We perform zero-shot cross-lingual polarity detection for each pair of the three languages. The point of cross-lingual classification is to train a model on labeled data from one language (\textit{source language}) and evaluate the model on data from another language (\textit{target language)}. The model must be able to transfer knowledge to the target language without any labeled data in the target language. For both languages in a pair, we train the model on data from one language and evaluate on data from the second language. The transformation of the semantic spaces is done in both directions, i.e., from the \textit{source} space to the \textit{target} space and vice versa. Since the French dataset is the largest one, for experiments where the source language is French, we use only French train data for training the model and as development data, we use French test data. For training models, where Czech is the source language, we use Czech train and test data parts of the CSFD dataset for training. We use the Czech dev part of the CSFD dataset as development data. In the case of the IMDB dataset,
we randomly selected 2,500 examples as development data and the rest is used to train the model. For the SST-2 dataset, we use train and dev parts for training and test part as the development data.

In every experiment, we evaluate and report results for test data part in the target language.
Only the test part of the data in the target language is used for evaluation, no other data from the target language are used to train the model.



\section{Experiments \& Results}
\label{sec:results}
We perform monolingual experiments, so we can compare the cross-lingual models with their monolingual equivalents. Secondly, we wanted to put our baselines and results into a context with the current SotA results. We use accuracy as the evaluation metric. We select the best models based on the results from the development data.
We repeat each experiment at least six times and report the arithmetic mean value with the 95\% confidence intervals.
The label distribution is nearly perfectly balanced in all datasets and thus, the resulting $F_{1}$ Macro score was in a vast majority of experiments almost identical to our accuracy and is therefore not reported in the paper. Thanks to this, we are able to compare our results with the existing work in \cite{priban-steinberger-2021-multilingual}.

\subsection{Monolingual Results}
Table~\ref{tab:monolingual-res} compares our proposed monolingual models with the current monolingual state-of-the-art (SotA) models.
We train both neural models with the existing fastText embeddings (rows CNN-F and LSTM-F) and with in-domain embeddings pre-trained by us (rows CNN and LSTM).
As we can see, the modern BERT-like models outperform all our baselines. For English, the difference is the largest among all other languages, especially for the SST-2 dataset.

For Czech and French, the difference between best baselines and SotA models is only $1.7\%$ and $1\%$, respectively. The difference between our baseline models trained on existing (rows CNN-F, LSTM-F) and our in-domain (rows CNN, LSTM) embeddings are at most $2.4\%$ and $2.5\%$ for CSFD and IMDB datasets, respectively. Based on the results, we can conclude that using custom pre-trained in-domain embeddings can slightly improve classification performance in monolingual settings, as one could expect. Our last observation is that our results are in general less competitive for English. This is most likely due to the fact that the current state-of-the-art approaches and models are generally more advanced in English, thus achieving better results.

\begin{table}[ht!]
\caption{Comparison of the monolingual results as accuracy (upper section) with the current monolingual SotA (bottom section).
The result with * was obtained on a custom data split. The results with $\ddagger$ are listed as $F_{1}$ score.} \label{tab:monolingual-res}
\begin{adjustbox}{width=\linewidth,center}

\setlength{\tabcolsep}{3pt}
\begin{tabular}{llllrcrrcrrc}
\toprule
\multicolumn{2}{c}{\textbf{CSFD} (Czech)} & \multicolumn{1}{c}{\textbf{}} & \multicolumn{2}{c}{\textbf{IMDB} (English)} &\multicolumn{1}{c}{} & \multicolumn{2}{c}{\textbf{SST-2} (English)} & \multicolumn{1}{c}{\textbf{}} & \multicolumn{2}{c}{\textbf{Allocine} (French)} & \multicolumn{1}{c}{\textbf{}} \\ \cmidrule{1-2} \cmidrule{4-5} \cmidrule{7-8} \cmidrule{10-11}

CNN (ours) & 93.9$^{\pm0.1}$  & &  & 91.8$^{\pm0.1}$ & & & 84.4$^{\pm0.6}$    & &  & 95.0$^{\pm0.1}$ &\\
CNN-F (ours) & 91.5$^{\pm0.2}$& &  & 89.3$^{\pm0.6}$ & & & 83.7$^{\pm0.2}$     &  & & 94.3$^{\pm0.1}$ &\\
LSTM (ours) & 94.3$^{\pm0.1}$& &  & 92.3$^{\pm0.4}$  & & & 84.5$^{\pm0.5}$     & & & 96.4$^{\pm0.1}$ &\\
LSTM-F (ours) & 92.1$^{\pm0.2}$& &  & 90.5$^{\pm0.9}$ & & & 84.3$^{\pm0.5}$    & & & 95.7$^{\pm0.1}$ &\\ \midrule
\multicolumn{11}{c}{\textbf{Current SotA}} \\ \midrule
LSTM \cite{priban-steinberger-2021-multilingual}$\ddagger$ & 91.8$^{\pm0.1}$ && \multicolumn{1}{l}{BON-Cos \cite{thongtan-phienthrakul-2019-sentiment}} & \multicolumn{1}{l}{97.4} && \multicolumn{1}{l}{RoB.\textsubscript{Smart} \cite{jiang-etal-2020-smart}} & \multicolumn{1}{l}{97.5} && \multicolumn{1}{l}{CamBERT \cite{allocine}} & \multicolumn{1}{l}{97.4} &\\
mBERT \cite{priban-steinberger-2021-multilingual}$\ddagger$ & 93.1$^{\pm0.3}$ && \multicolumn{1}{l}{XLNet \cite{XLNet}} & \multicolumn{1}{l}{96.2} && \multicolumn{1}{l}{T5-11B \cite{t5-11b}} & \multicolumn{1}{l}{97.5} && \multicolumn{1}{l}{CNN \cite{allocine}} & \multicolumn{1}{l}{93.7} &\\
XLM-R\textsubscript{Large} \cite{priban-steinberger-2021-multilingual}$\ddagger$ & 96.0$^{\pm0.0}$ && \multicolumn{1}{l}{BERT\textsubscript{ITPT} \cite{sun2019fine}} & \multicolumn{1}{l}{95.8} && \multicolumn{1}{l}{XLNet \cite{XLNet}} & \multicolumn{1}{l}{97.0} &&
& \multicolumn{1}{l}{93.0}& \\
BERT\textsubscript{Distilled}\cite{kybernetika-bert}* & \multicolumn{1}{l}{93.8}& & \multicolumn{1}{l}{oh-LSTM \cite{oh-lstm}} & \multicolumn{1}{l}{94.1} & & &&  &&  &
\\ \bottomrule  

\end{tabular}
\end{adjustbox}

\end{table}

\subsection{Cross-Lingual Results}
We report our cross-lingual results for all three pairs of languages in Tables~\ref{tab:cross-lingual-en-cs}, \ref{tab:cross-lingual-en-fr} and \ref{tab:cross-lingual-fr-cs}. In each table, we present results trained with \textit{in-domain} embeddings pre-trained by us and results for existing \textit{fastText} embeddings, separated by the slash character. These pairs of results were always obtained by models trained with the same hyper-parameters (learning rate and the number of epochs). We report the results of experiments where the semantic spaces were transformed in both directions\footnote{For example, the column labeled as \textbf{EN}-\textbf{s} $\Rightarrow$ \textbf{CS}-\textbf{t} means that English space was transformed into Czech space. English is the source language (-s suffix) and Czech is the target language (-t suffix), in other words, the English dataset is used for training and Czech for the evaluation.}. For easier comparison, we also include the monolingual results of our models from Table \ref{tab:monolingual-res}. The pairs where in-domain embeddings are better than the existing fastText embeddings have a gray background.
The best results in absolute values are underlined and the results that overlap with the confidence interval of the best result are bold,
we mark this separately in each column. As we mentioned, we trained the models for at most five epochs with constant learning rate or linear learning rate decay with learning rates of 1e-3 or 1e-4\footnote{We provide the details of the used hyper-parameters at our GitHub repository.}.

\par Our main observation is that in-domain embeddings significantly improve the results for the CLSA task (gray background in the Tables). The improvement is in some cases, even more than 10\%. This statement is certainly true for the models trained on English and evaluated on Czech and French (with some minor exceptions). For the models evaluated on English, the improvement is not so noticeable, but in most cases, it is also valid. We can observe an analogical improvement in the monolingual results, but for these, the improvement is at most 2.5\%.



\setlength\fboxsep{0pt}
\begin{table}[ht!]
\caption{Cross-lingual accuracy results for English and Czech language pair.} \label{tab:cross-lingual-en-cs}

\begin{adjustbox}{width=\linewidth,center}
\begin{tabular}{llccccclclclc} \toprule 
 \multicolumn{1}{l}{} &  & \multicolumn{5}{c}{Evaluated on \textbf{Czech}} &  & \multicolumn{5}{c}{Evaluated on \textbf{English}} \\ \cmidrule{3-7} \cmidrule{9-13}
 &  &  &  & \textbf{EN}-\textbf{s} $\Rightarrow$ \textbf{CS}-\textbf{t} &  & \textbf{CS}-\textbf{t} $\Rightarrow$ \textbf{EN}-\textbf{s} &  &  &  & \multicolumn{1}{c}{\textbf{CS}-\textbf{s} $\Rightarrow$ \textbf{EN}-\textbf{t}} &  & \textbf{EN}-\textbf{t} $\Rightarrow$ \textbf{CS}-\textbf{s} \\ \cline{5-5} \cline{7-7} \cline{11-11} \cline{13-13}
\textbf{Dataset} & \multicolumn{1}{c}{\textbf{Method}} & \multicolumn{1}{c}{\textbf{Monoling.}} &  & \multicolumn{1}{c}{in-domain/fastText} &  & in-domain/fastText &  & \multicolumn{1}{c}{\textbf{Monoling.}}  &  & \multicolumn{1}{l}{in-domain/fastText} &  & in-domain/fastText \\ \midrule
 \multicolumn{13}{c}{\textbf{CNN}} \\ \cdashlinelr{2-13}

\multirow{12}{*}{\shortstack{IMDB \\ CSFD}} & MSE & \multirow{5}{*}{\shortstack{93.9/ \\91.5\phantom{*}}} &  & \colorbox{gray!20}{\textbf{88.2}$^{\pm0.3}$/75.7$^{\pm1.5}$} &  & \colorbox{gray!20}{72.3$^{\pm2.2}$/69.0$^{\pm2.0}$} &  & \multirow{5}{*}{\shortstack{91.8/ \\89.2\phantom{*}}} &  & \colorbox{gray!20}{77.5$^{\pm1.5}$/67.1$^{\pm1.9}$} &  & 53.8$^{\pm2.0}$/67.1$^{\pm1.4}$ \\
 & Orto &  &  & \colorbox{gray!20}{\underline{\textbf{88.5}}$^{\pm0.1}$/78.9$^{\pm0.9}$} &  & \colorbox{gray!20}{87.4$^{\pm0.9}$/72.5$^{\pm1.4}$} &  &  &  & \colorbox{gray!20}{83.8$^{\pm0.1}$/76.8$^{\pm0.3}$} &  & \colorbox{gray!20}{81.3$^{\pm0.3}$/79.3$^{\pm0.9}$} \\
 & CCA &  &  & \colorbox{gray!20}{\textbf{88.4}$^{\pm0.1}$/76.2$^{\pm1.2}$} &  & \colorbox{gray!20}{87.4$^{\pm0.4}$/79.5$^{\pm0.6}$} &  &  &  & \colorbox{gray!20}{83.9$^{\pm0.1}$/75.0$^{\pm0.6}$} &  & \colorbox{gray!20}{79.6$^{\pm0.6}$/67.2$^{\pm4.4}$} \\
 & Rank &  &  & \colorbox{gray!20}{85.7$^{\pm0.3}$/78.9$^{\pm0.9}$} &  & \colorbox{gray!20}{88.0$^{\pm0.8}$/76.7$^{\pm0.8}$} &  &  &  & \colorbox{gray!20}{83.2$^{\pm0.2}$/76.1$^{\pm0.7}$} &  & \colorbox{gray!20}{\textbf{82.0}$^{\pm1.0}$/74.4$^{\pm1.9}$} \\
 & Or-Ra &  &  & \colorbox{gray!20}{83.3$^{\pm0.7}$/76.9$^{\pm1.8}$} &  & \colorbox{gray!20}{\underline{\textbf{89.2}}$^{\pm0.1}$/79.2$^{\pm1.0}$} &  &  &  & \colorbox{gray!20}{79.6$^{\pm0.5}$/78.4$^{\pm0.6}$} &  & \colorbox{gray!20}{\textbf{82.3}$^{\pm0.4}$/75.0$^{\pm1.0}$} \\ 
 
 \multicolumn{13}{c}{\textbf{LSTM}} \\  \cdashlinelr{2-13}
 & MSE & \multirow{5}{*}{\shortstack{94.3/ \\92.1\phantom{*}}} &  & \colorbox{gray!20}{84.9$^{\pm0.6}$/80.6$^{\pm1.3}$} &  & \colorbox{gray!20}{83.4$^{\pm2.0}$/79.7$^{\pm2.1}$} &  & \multirow{5}{*}{\shortstack{92.3/ \\90.5\phantom{*}}} &  & \colorbox{gray!20}{\underline{\textbf{84.8}}$^{\pm0.6}$/\textbf{83.4}$^{\pm1.0}$} &  & 51.2$^{\pm2.2}$/62.9$^{\pm3.7}$ \\
 & Orto &  &  & \colorbox{gray!20}{87.1$^{\pm0.3}$/81.5$^{\pm1.3}$} &  & \colorbox{gray!20}{87.7$^{\pm0.7}$/82.2$^{\pm0.8}$} &  &  &  & 73.6$^{\pm1.4}$/79.8$^{\pm1.7}$ &  & 68.2$^{\pm1.5}$/\textbf{83.7}$^{\pm0.7}$ \\
 & CCA &  &  & \colorbox{gray!20}{85.9$^{\pm1.7}$/81.3$^{\pm1.5}$} &  & \colorbox{gray!20}{87.4$^{\pm0.3}$/82.6$^{\pm0.4}$} &  &  &  & \colorbox{gray!20}{\textbf{82.7}$^{\pm2.9}$/77.7$^{\pm2.9}$} &  & \textbf{81.8}$^{\pm1.6}$/\textbf{82.6}$^{\pm0.7}$ \\
 & Rank &  &  & \colorbox{gray!20}{82.5$^{\pm2.4}$/76.9$^{\pm1.4}$} &  & \colorbox{gray!20}{85.1$^{\pm1.4}$/80.9$^{\pm2.9}$} &  &  &  & 56.3$^{\pm2.4}$/\textbf{82.9}$^{\pm1.9}$ &  & \colorbox{gray!20}{\underline{\textbf{83.8}}$^{\pm0.9}$/\textbf{83.5}$^{\pm0.5}$} \\
 & Or-Ra &  &  & \colorbox{gray!20}{86.2$^{\pm0.5}$/73.2$^{\pm1.4}$} &  & \colorbox{gray!20}{86.7$^{\pm0.7}$/82.9$^{\pm2.1}$} &  &  &  & 67.9$^{\pm3.2}$/\textbf{82.8}$^{\pm2.0}$ &  & \colorbox{gray!20}{\textbf{83.6}$^{\pm1.2}$/\textbf{83.1}$^{\pm0.4}$} \\ \midrule 
 
  \multicolumn{13}{c}{\textbf{CNN}} \\ \cdashlinelr{2-13}
\multirow{12}{*}{\shortstack{SST-2 \\ CSFD}} & MSE & \multirow{5}{*}{\shortstack{93.9/ \\91.5\phantom{*}}} &  & \colorbox{gray!20}{\textbf{84.1}$^{\pm2.1}$/55.7$^{\pm3.5}$} &  & \colorbox{gray!20}{\underline{\textbf{86.0}}$^{\pm1.4}$/78.1$^{\pm1.0}$} &  & \multirow{5}{*}{\shortstack{84.4/ \\83.6\phantom{*}}} &  & 72.8$^{\pm0.3}$/73.1$^{\pm0.3}$ &  & 50.5$^{\pm1.5}$/60.1$^{\pm3.4}$ \\
 & Orto &  &  & \colorbox{gray!20}{77.2$^{\pm1.1}$/50.9$^{\pm2.2}$} &  & \colorbox{gray!20}{81.8$^{\pm1.7}$/74.9$^{\pm0.9}$} &  &  &  & \colorbox{gray!20}{\textbf{77.8}$^{\pm0.2}$/76.0$^{\pm0.2}$} &  & \colorbox{gray!20}{75.3$^{\pm0.5}$/74.3$^{\pm0.9}$} \\
 & CCA &  &  & \colorbox{gray!20}{\textbf{83.9}$^{\pm0.7}$/51.2$^{\pm1.1}$} &  & \colorbox{gray!20}{83.1$^{\pm0.7}$/76.6$^{\pm0.9}$} &  &  &  & \colorbox{gray!20}{77.2$^{\pm0.2}$/75.0$^{\pm0.3}$} &  & \colorbox{gray!20}{72.4$^{\pm0.2}$/72.1$^{\pm0.3}$} \\
 & Rank &  &  & \colorbox{gray!20}{\underline{\textbf{85.2}}$^{\pm0.6}$/55.8$^{\pm2.7}$} &  & \colorbox{gray!20}{83.2$^{\pm1.6}$/75.6$^{\pm0.4}$} &  &  &  & \colorbox{gray!20}{77.0$^{\pm0.5}$/73.2$^{\pm0.2}$} &  & 77.4$^{\pm0.4}$/75.2$^{\pm0.4}$ \\
 & Or-Ra &  &  & \colorbox{gray!20}{80.1$^{\pm1.5}$/55.6$^{\pm3.6}$} &  & \colorbox{gray!20}{82.6$^{\pm1.4}$/76.9$^{\pm0.3}$} &  &  &  & \colorbox{gray!20}{76.2$^{\pm0.4}$/75.5$^{\pm0.4}$} &  & \colorbox{gray!20}{77.4$^{\pm0.3}$/77.2$^{\pm0.3}$} \\

\multicolumn{13}{c}{\textbf{LSTM}} \\   \cdashlinelr{2-13}

 & MSE & \multirow{5}{*}{\shortstack{94.3/ \\92.1\phantom{*}}} &  & \colorbox{gray!20}{81.1$^{\pm1.9}$/76.4$^{\pm2.9}$} &  & \colorbox{gray!20}{82.0$^{\pm2.3}$/69.5$^{\pm2.6}$} &  & \multirow{5}{*}{\shortstack{84.5/ \\84.3\phantom{*}}} &  & 76.1$^{\pm0.4}$/\underline{{\textbf{78.4}}}$^{\pm0.4}$ &  & 68.5$^{\pm0.8}$/73.0$^{\pm2.2}$ \\
 & Orto &  &  & \colorbox{gray!20}{80.4$^{\pm2.1}$/75.5$^{\pm1.3}$} &  & \colorbox{gray!20}{76.4$^{\pm2.3}$/74.8$^{\pm1.3}$} &  &  &  & 72.6$^{\pm1.6}$/\textbf{78.4}$^{\pm0.5}$ &  & 72.9$^{\pm0.5}$/\textbf{79.2}$^{\pm1.0}$ \\
 & CCA &  &  & \colorbox{gray!20}{83.0$^{\pm1.4}$/72.7$^{\pm2.1}$} &  & \colorbox{gray!20}{82.5$^{\pm0.8}$/72.9$^{\pm2.2}$} &  &  &  & \textbf{76.5}$^{\pm2.0}$/\textbf{76.9}$^{\pm1.7}$ &  & 73.9$^{\pm1.8}$/75.9$^{\pm1.0}$ \\
 & Rank &  &  & \colorbox{gray!20}{83.1$^{\pm1.3}$/74.6$^{\pm2.2}$} &  & 75.5$^{\pm2.1}$/77.8$^{\pm2.0}$ &  &  &  & 70.8$^{\pm1.7}$/77.1$^{\pm0.8}$ &  & 77.5$^{\pm1.5}$/\textbf{79.1}$^{\pm0.6}$ \\
 & Or-Ra &  &  & \colorbox{gray!20}{83.0$^{\pm0.7}$/73.8$^{\pm1.7}$} &  & \colorbox{gray!20}{82.2$^{\pm1.9}$/78.3$^{\pm2.4}$} &  &  &  & 74.7$^{\pm1.8}$/\textbf{76.1}$^{\pm2.1}$ &  & 75.9$^{\pm2.5}$/\underline{\textbf{79.5}}$^{\pm0.4}$ \\ \bottomrule

\end{tabular}
\end{adjustbox}
\end{table}

\begin{table}[ht!]
\caption{Cross-lingual accuracy results for English and French language pair.} \label{tab:cross-lingual-en-fr}
\begin{adjustbox}{width=\linewidth,center}
\begin{tabular}{llclclclclllc} \toprule
\multicolumn{1}{l}{} &  & \multicolumn{5}{c}{Evaluated on \textbf{French}} &  & \multicolumn{5}{c}{Evaluated on \textbf{English}} \\ \cmidrule{3-7} \cmidrule{9-13}
\multicolumn{1}{l}{} &  & \multicolumn{1}{l}{} &  & \textbf{EN}-\textbf{s} $\Rightarrow$ \textbf{FR}-\textbf{t} &  & \textbf{FR}-\textbf{t} $\Rightarrow$ \textbf{EN}-\textbf{s} &  & \multicolumn{1}{l}{} &  & \textbf{FR}-\textbf{s} $\Rightarrow$ \textbf{EN}-\textbf{t} &  & \textbf{EN}-\textbf{t} $\Rightarrow$ \textbf{FR}-\textbf{s} \\ \cline{5-5} \cline{7-7} \cline{11-11} \cline{13-13}
\textbf{Dataset} & \textbf{Method} & \textbf{Monoling.} &  & in-domain/fastText &  & in-domain/fastText &  & \textbf{Monoling.} &  & \multicolumn{1}{c}{in-domain/fastText} &  & in-domain/fastText \\ \midrule
 \multicolumn{13}{c}{\textbf{CNN}} \\ \cdashlinelr{2-13}
\multirow{12}{*}{\shortstack{IMDB \\ Allocine}} & MSE & \multirow{5}{*}{\shortstack{95.0/ \\94.3\phantom{*}}} &  & \colorbox{gray!20}{86.5$^{\pm0.2}$/81.3$^{\pm0.2}$} &  & 63.2$^{\pm9.1}$/79.9$^{\pm0.7}$ &  & \multirow{5}{*}{\shortstack{91.8/ \\89.2\phantom{*}}} &  & \colorbox{gray!20}{\underline{\textbf{86.2}}$^{\pm0.1}$/78.2$^{\pm0.1}$} &  & 55.1$^{\pm4.4}$/72.8$^{\pm1.2}$ \\
 & Orto &  &  & \colorbox{gray!20}{90.4$^{\pm0.0}$/81.0$^{\pm0.6}$} &  & \colorbox{gray!20}{89.0$^{\pm0.2}$/81.3$^{\pm0.3}$} &  &  &  & \colorbox{gray!20}{86.0$^{\pm0.0}$/81.3$^{\pm0.4}$} &  & \colorbox{gray!20}{\textbf{87.0}$^{\pm0.0}$/81.0$^{\pm0.6}$} \\
 & CCA &  &  & \colorbox{gray!20}{89.9$^{\pm0.1}$/81.0$^{\pm0.5}$} &  & \colorbox{gray!20}{89.0$^{\pm0.1}$/81.9$^{\pm0.0}$} &  &  &  & \colorbox{gray!20}{84.6$^{\pm0.2}$/80.2$^{\pm0.4}$} &  & \colorbox{gray!20}{85.3$^{\pm0.2}$/79.2$^{\pm0.4}$} \\
 & Rank &  &  & \colorbox{gray!20}{88.2$^{\pm0.9}$/77.1$^{\pm2.6}$} &  & \colorbox{gray!20}{88.7$^{\pm0.0}$/80.9$^{\pm0.4}$} &  &  &  & \colorbox{gray!20}{83.6$^{\pm0.1}$/74.5$^{\pm0.7}$} &  & \colorbox{gray!20}{85.3$^{\pm0.5}$/74.9$^{\pm0.9}$} \\
 & Or-Ra &  &  & \colorbox{gray!20}{89.2$^{\pm0.2}$/75.9$^{\pm0.8}$} &  & \colorbox{gray!20}{\underline{\textbf{89.4}}$^{\pm0.0}$/80.8$^{\pm0.6}$} &  &  &  & \colorbox{gray!20}{81.0$^{\pm0.9}$/78.3$^{\pm1.5}$} &  & \colorbox{gray!20}{86.3$^{\pm0.1}$/76.2$^{\pm0.8}$} \\ 
 
 \multicolumn{13}{c}{\textbf{LSTM}} \\  \cdashlinelr{2-13}
 
   & MSE & \multirow{5}{*}{\shortstack{96.4/ \\95.7\phantom{*}}} &  & \colorbox{gray!20}{84.9$^{\pm1.2}$/80.2$^{\pm9.0}$} &  & 68.7$^{\pm9.7}$/68.7$^{\pm9.6}$ &  & \multirow{5}{*}{\shortstack{92.3/ \\90.5\phantom{*}}} &  & 81.7$^{\pm1.0}$/81.9$^{\pm0.6}$ &  & \colorbox{gray!20}{79.3$^{\pm2.9}$/78.4$^{\pm1.1}$} \\
 & Orto &  &  & \colorbox{gray!20}{\textbf{91.2}$^{\pm0.3}$/81.2$^{\pm8.2}$} &  & \colorbox{gray!20}{86.9$^{\pm4.8}$/82.4$^{\pm4.1}$} &  &  &  & 81.6$^{\pm1.2}$/83.0$^{\pm0.5}$ &  & \colorbox{gray!20}{83.2$^{\pm2.2}$/80.3$^{\pm2.1}$} \\
 & CCA &  &  & \colorbox{gray!20}{\underline{\textbf{91.7}}$^{\pm0.2}$/85.0$^{\pm1.7}$} &  & \colorbox{gray!20}{87.8$^{\pm2.2}$/85.9$^{\pm1.1}$} &  &  &  & \colorbox{gray!20}{\textbf{85.2}$^{\pm0.9}$/82.2$^{\pm0.1}$} &  & \colorbox{gray!20}{\textbf{86.8}$^{\pm0.3}$/71.5$^{\pm2.4}$} \\
 & Rank &  &  & \colorbox{gray!20}{88.3$^{\pm1.6}$/87.2$^{\pm1.0}$} &  & \colorbox{gray!20}{88.2$^{\pm3.2}$/82.9$^{\pm3.6}$} &  &  &  & 56.7$^{\pm4.6}$/81.1$^{\pm0.9}$ &  & \colorbox{gray!20}{85.2$^{\pm0.6}$/81.2$^{\pm0.5}$} \\
 & Or-Ra &  &  & \colorbox{gray!20}{86.3$^{\pm3.4}$/70.9$^{\pm9.0}$} &  & \colorbox{gray!20}{\textbf{89.1}$^{\pm1.2}$/86.8$^{\pm0.7}$} &  &  &  & 56.4$^{\pm3.1}$/81.5$^{\pm1.5}$ &  & \colorbox{gray!20}{\underline{\textbf{87.2}}$^{\pm0.2}$/79.9$^{\pm1.8}$} \\ \midrule 
 
  \multicolumn{13}{c}{\textbf{CNN}} \\ \cdashlinelr{2-13}
 
\multirow{12}{*}{\shortstack{SST-2 \\ Allocine}} & MSE & \multirow{5}{*}{\shortstack{95.0/ \\94.3\phantom{*}}} &  & \colorbox{gray!20}{86.7$^{\pm0.8}$/67.9$^{\pm2.1}$} &  & \colorbox{gray!20}{\underline{\textbf{84.5}}$^{\pm0.2}$/68.4$^{\pm0.9}$} &  & \multirow{5}{*}{\shortstack{84.4/ \\83.6\phantom{*}}} &  & \colorbox{gray!20}{\textbf{79.6}$^{\pm0.1}$/79.2$^{\pm0.3}$} &  & 50.8$^{\pm1.3}$/72.8$^{\pm1.8}$ \\
 & Orto &  &  & \colorbox{gray!20}{85.9$^{\pm0.5}$/65.7$^{\pm2.9}$} &  & \colorbox{gray!20}{81.0$^{\pm0.7}$/69.1$^{\pm2.4}$} &  &  &  & 78.9$^{\pm0.3}$/\underline{\textbf{80.0}}$^{\pm0.3}$ &  & \colorbox{gray!20}{80.1$^{\pm0.3}$/79.5$^{\pm0.2}$} \\
 & CCA &  &  & \colorbox{gray!20}{\underline{\textbf{87.8}}$^{\pm0.1}$/61.7$^{\pm2.4}$} &  & \colorbox{gray!20}{\textbf{84.2}$^{\pm0.6}$/66.4$^{\pm1.1}$} &  &  &  & 77.7$^{\pm0.3}$/78.9$^{\pm0.2}$ &  & \colorbox{gray!20}{79.2$^{\pm0.3}$/78.2$^{\pm0.3}$} \\
 & Rank &  &  & \colorbox{gray!20}{85.9$^{\pm0.4}$/68.1$^{\pm1.8}$} &  & \colorbox{gray!20}{81.9$^{\pm1.2}$/68.1$^{\pm1.1}$} &  &  &  & \colorbox{gray!20}{75.4$^{\pm1.0}$/74.4$^{\pm1.2}$} &  & \colorbox{gray!20}{81.8$^{\pm0.2}$/78.8$^{\pm0.2}$} \\
 & Or-Ra &  &  & \colorbox{gray!20}{83.8$^{\pm0.3}$/72.1$^{\pm2.0}$} &  & \colorbox{gray!20}{\textbf{83.9}$^{\pm0.9}$/70.7$^{\pm0.9}$} &  &  &  & 73.4$^{\pm1.3}$/77.8$^{\pm0.6}$ &  & \colorbox{gray!20}{80.9$^{\pm0.2}$/79.1$^{\pm0.3}$} \\ 
\multicolumn{13}{c}{\textbf{LSTM}} \\ \cdashlinelr{2-13}

 & MSE & \multirow{5}{*}{\shortstack{96.4/ \\95.7\phantom{*}}} &  & \colorbox{gray!20}{78.6$^{\pm3.7}$/77.6$^{\pm1.7}$} &  & \colorbox{gray!20}{81.9$^{\pm0.9}$/71.6$^{\pm9.2}$} &  & \multirow{5}{*}{\shortstack{84.5/ \\84.3\phantom{*}}} &  & \colorbox{gray!20}{79.1$^{\pm0.1}$/78.8$^{\pm0.7}$} &  & 76.2$^{\pm0.8}$/76.3$^{\pm0.6}$ \\
 & Orto &  &  & \colorbox{gray!20}{84.7$^{\pm0.5}$/76.2$^{\pm4.6}$} &  & \colorbox{gray!20}{81.1$^{\pm2.8}$/\textbf{78.1}$^{\pm6.2}$} &  &  &  & \textbf{79.2}$^{\pm0.6}$/79.4$^{\pm0.1}$ &  & \colorbox{gray!20}{\textbf{81.8}$^{\pm0.3}$/78.9$^{\pm0.3}$} \\
 & CCA &  &  & \colorbox{gray!20}{85.3$^{\pm0.8}$/79.6$^{\pm1.1}$} &  & \colorbox{gray!20}{\textbf{81.8}$^{\pm4.9}$/78.4$^{\pm1.3}$} &  &  &  & \colorbox{gray!20}{\textbf{79.9}$^{\pm0.3}$/78.6$^{\pm0.6}$} &  & \colorbox{gray!20}{80.7$^{\pm0.4}$/78.1$^{\pm0.4}$} \\
 & Rank &  &  & \colorbox{gray!20}{85.3$^{\pm1.7}$/77.3$^{\pm3.1}$} &  & \colorbox{gray!20}{81.9$^{\pm1.1}$/79.5$^{\pm1.1}$} &  &  &  & 69.8$^{\pm3.1}$/77.4$^{\pm0.2}$ &  & \colorbox{gray!20}{\underline{\textbf{82.5}}$^{\pm0.4}$/79.0$^{\pm0.4}$} \\
 & Or-Ra &  &  & \colorbox{gray!20}{84.7$^{\pm1.9}$/75.7$^{\pm2.7}$} &  & \colorbox{gray!20}{\textbf{82.3}$^{\pm5.0}$/76.6$^{\pm3.4}$} &  &  &  & 72.7$^{\pm2.1}$/78.5$^{\pm0.6}$ &  & \colorbox{gray!20}{\textbf{81.7}$^{\pm0.4}$/79.6$^{\pm0.4}$} \\
 
 \bottomrule
 
\end{tabular}
\end{adjustbox}
\end{table}

\subsection{Comparison with Existing Works}
\par Table \ref{tab:cross-lingual-comparison-other-works} compares our best results with  related work. This table shows that the proposed approach based on linear transformations is competitive with the current BERT-like models. There is an exception for English results obtained by the XLM-R\textsubscript{Large} model, which has a huge number of parameters (559M). We did not outperform the largest XLM-R\textsubscript{Large} model, but for example, for the CSFD dataset, we beat three BERT-like models that are much larger (in terms of a number of parameters) than our CNN and LSTM models. In the case of the mBERT and XLM models, the difference is very significant. The results in Table \ref{tab:cross-lingual-comparison-other-works} are shown as Macro $F_1$, but we have to note that our Macro $F_1$ results are identical to the accuracy.

\begin{table}[ht!]
\caption{Comparison of cross-lingual Macro $F_1$ results with other works. French result with * symbol is shown as accuracy.} \label{tab:cross-lingual-comparison-other-works}
\centering
\begin{adjustbox}{width=0.55\linewidth,center}
\begin{tabular}{lccc} \toprule
 & IMDB & CSFD & Allocine \\ \midrule
XLM-R\textsubscript{Base} \cite{priban-steinberger-2021-multilingual} & 89.5$^{\pm0.2}$  & 88.0$^{\pm0.3}$ &  \\
XLM-R\textsubscript{Large} \cite{priban-steinberger-2021-multilingual} & 94.0$^{\pm0.1}$  & 91.6$^{\pm0.1}$  &  \\
XLM \cite{priban-steinberger-2021-multilingual} & 78.2$^{\pm0.5}$  & 75.4$^{\pm0.3}$  &  \\
mBERT \cite{priban-steinberger-2021-multilingual} &  & 76.3$^{\pm1.1}$  &  \\
G/F-A \cite{dong2018cross}* &  &  & \multicolumn{1}{c}{93.0}\phantom{***} \\
 \cdashlinelr{1-4}
\multirow{2}{*}{Our best} & \multicolumn{1}{c}{\multirow{2}{*}{\shortstack{87.2$^{\pm0.2}$ \\ (EN-t $\Rightarrow$ FR-s)}}} & \multirow{2}{*}{\shortstack{89.2$^{\pm0.1}$ \\ (CS-t $\Rightarrow$ EN-s)}} & \multirow{2}{*}{\shortstack{91.7$^{\pm0.2}$ \\ (EN-s $\Rightarrow$ FR-t)}}

\\
 \\ \bottomrule
\end{tabular}
\end{adjustbox}
\end{table}

\begin{table}[ht!]
\catcode`\-=12
\caption{Cross-lingual accuracy results for French and Czech language pair.} \label{tab:cross-lingual-fr-cs}

\begin{adjustbox}{width=\linewidth,center}
\begin{tabular}{llclclclclllc} \toprule
\multicolumn{1}{l}{} &  & \multicolumn{5}{c}{Evaluated on \textbf{Czech}} &  & \multicolumn{5}{c}{Evaluated on \textbf{French}} \\ \cmidrule{3-7} \cmidrule{9-13}
\multicolumn{1}{l}{} &  & \multicolumn{1}{l}{} &  & \textbf{FR}-\textbf{s} $\Rightarrow$ \textbf{CS}-\textbf{t} &  & \textbf{CS}-\textbf{t} $\Rightarrow$ \textbf{FR}-\textbf{s} &  & \multicolumn{1}{l}{} &  & \textbf{CS}-\textbf{s} $\Rightarrow$ \textbf{FR}-\textbf{t} &  & \textbf{FR}-\textbf{t} $\Rightarrow$ \textbf{CS}-\textbf{s} \\ \cline{5-5} \cline{7-7} \cline{11-11} \cline{13-13}
\textbf{Dataset} & \textbf{Method} & \textbf{Monoling.} &  & in-domain/fastText &  & in-domain/fastText &  & \textbf{Monoling.} &  & \multicolumn{1}{c}{in-domain/fastText} &  & in-domain/fastText \\ \midrule
 \multicolumn{13}{c}{\textbf{CNN}} \\ \cdashlinelr{2-13}
\multirow{5}{*}{\shortstack{Allocine\\ CSFD}} & MSE & \multirow{5}{*}{\shortstack{93.9/ \\91.5\phantom{*}}} &  & \colorbox{gray!20}{84.4$^{\pm0.2}$/75.2$^{\pm1.2}$} &  & 56.0$^{\pm3.1}$/68.5$^{\pm3.9}$ &  & \multirow{5}{*}{\shortstack{95.0/ \\94.3\phantom{*}}} &  & \colorbox{gray!20}{74.3$^{\pm0.7}$/65.6$^{\pm1.0}$} &  & 52.7$^{\pm0.4}$/63.8$^{\pm2.8}$ \\
 & Orto &  &  & \colorbox{gray!20}{85.9$^{\pm0.3}$/77.5$^{\pm0.5}$} &  & \colorbox{gray!20}{86.0$^{\pm0.3}$/78.0$^{\pm0.3}$} &  &  &  & \colorbox{gray!20}{\textbf{84.6}$^{\pm0.2}$/80.8$^{\pm0.2}$} &  & \colorbox{gray!20}{84.0$^{\pm0.3}$/78.4$^{\pm0.5}$} \\
 & CCA &  &  & \colorbox{gray!20}{83.7$^{\pm0.3}$/75.9$^{\pm0.4}$} &  & \colorbox{gray!20}{82.7$^{\pm0.6}$/71.8$^{\pm0.5}$} &  &  &  & \colorbox{gray!20}{\underline{\textbf{84.7}}$^{\pm0.3}$/79.8$^{\pm0.3}$} &  & \colorbox{gray!20}{76.9$^{\pm0.5}$/73.7$^{\pm0.6}$} \\
 & Rank &  &  & \colorbox{gray!20}{81.7$^{\pm1.0}$/75.1$^{\pm1.3}$} &  & \colorbox{gray!20}{86.2$^{\pm0.3}$/69.2$^{\pm0.2}$} &  &  &  & \colorbox{gray!20}{82.4$^{\pm0.8}$/78.5$^{\pm0.2}$} &  & \colorbox{gray!20}{\textbf{84.6}$^{\pm0.1}$/68.9$^{\pm1.3}$} \\
 & Or-Ra &  &  & \colorbox{gray!20}{82.7$^{\pm0.8}$/72.6$^{\pm1.6}$} &  & \colorbox{gray!20}{87.0$^{\pm0.1}$/74.3$^{\pm0.9}$} &  &  &  & \colorbox{gray!20}{75.9$^{\pm1.4}$/71.8$^{\pm2.8}$} &  & \colorbox{gray!20}{\textbf{85.3}$^{\pm0.2}$/80.3$^{\pm0.2}$}   \\ 
\multicolumn{13}{c}{\textbf{LSTM}} \\ \cdashlinelr{2-13}
 
 \multirow{5}{*}{\shortstack{Allocine\\ CSFD}} & MSE & \multirow{5}{*}{\shortstack{94.3/ \\92.1\phantom{*}}} &  & \colorbox{gray!20}{85.3$^{\pm0.6}$/81.5$^{\pm1.1}$} &  & \colorbox{gray!20}{84.1$^{\pm4.1}$/76.5$^{\pm2.8}$} &  & \multirow{5}{*}{\shortstack{96.4/ \\95.7\phantom{*}}} &  & \colorbox{gray!20}{81.7$^{\pm2.1}$/76.6$^{\pm1.9}$} &  & 52.5$^{\pm2.5}$/62.2$^{\pm5.7}$ \\
 & Orto &  &  & \colorbox{gray!20}{\underline{\textbf{87.6}}$^{\pm0.6}$/80.2$^{\pm0.6}$} &  & \colorbox{gray!20}{\underline{\textbf{88.0}}$^{\pm0.7}$/81.5$^{\pm0.7}$} &  &  &  & \colorbox{gray!20}{71.8$^{\pm0.9}$/70.7$^{\pm4.5}$} &  & 68.9$^{\pm1.1}$/69.9$^{\pm5.3}$ \\
 & CCA &  &  & \colorbox{gray!20}{\textbf{87.4}$^{\pm0.4}$/79.3$^{\pm1.2}$} &  & \colorbox{gray!20}{\textbf{87.3}$^{\pm0.5}$/79.3$^{\pm1.0}$} &  &  &  & \colorbox{gray!20}{76.5$^{\pm2.9}$/72.5$^{\pm4.0}$} &  & 64.0$^{\pm1.4}$/72.5$^{\pm3.3}$ \\
 & Rank &  &  & 76.6$^{\pm5.9}$/81.2$^{\pm1.1}$ &  & \colorbox{gray!20}{86.4$^{\pm0.7}$/76.1$^{\pm1.3}$} &  &  &  & 69.2$^{\pm7.3}$/77.1$^{\pm2.9}$ &  & \colorbox{gray!20}{\underline{\textbf{85.4}}$^{\pm0.8}$/78.5$^{\pm1.0}$} \\
 & Or-Ra &  &  & \colorbox{gray!20}{84.0$^{\pm2.2}$/78.7$^{\pm3.4}$} &  & \colorbox{gray!20}{\textbf{87.6}$^{\pm0.6}$/81.0$^{\pm0.9}$} &  &  &  & \colorbox{gray!20}{78.4$^{\pm5.8}$/58.1$^{\pm4.6}$} &  & \colorbox{gray!20}{83.3$^{\pm1.1}$/82.7$^{\pm0.7}$} \\ \bottomrule

\end{tabular}
\end{adjustbox}
\end{table}

\subsection{Discussion}
From our perception, during the experiments and from the presented Tables, we consider the CCA and Orthogonal transformations to be the most stable in terms of performance. These two methods usually obtain very comparable (if not the best) results across all experiments, unlike the other methods. The other methods (MSE, Rank Or-Ra) tend to fail in some settings and their performance decreases significantly. For example, the performance of the MSE method often drops by a large margin for experiments where embeddings for the target language is mapped into the source language (the \textbf{FR}-\textbf{t} $\Rightarrow$ \textbf{EN}-\textbf{s} columns).

\par We found no systematic differences in performance between the LSTM and CNN models. We recognized that the transformations sometimes fail and cause poor performance of the overall model. In future work, we want to focus on these particular incidents and we would like to improve the stability of the transformations, for example, by normalizing the semantic spaces before and after the transformation.




\section{Conclusion}
In this paper, we studied the task of cross-lingual sentiment analysis for Czech, English and French. We performed zero-shot cross-lingual classification on four datasets with linear transformations in combination with LSTM and CNN based classifiers. We demonstrate that pre-trained in-domain embeddings can significantly improve the cross-lingual classification, in some cases even by more than 10\%. We show that the approaches based on linear transformations are, to some extent, competitive with the multilingual BERT-like models. We provide all the presented resources, including word embeddings, dictionaries and source codes, freely for research purposes on our GitHub\footref{fnlabel}.

\section*{Acknowledgments}
This work has been partly supported by ERDF ”Research and Development of Intelligent Components of Advanced Technologies for the Pilsen Metropolitan Area (InteCom)” (no.: CZ.02.1.01/0.0/0.0/17 048/0007267); and by Grant No. SGS-2022-016 Advanced methods of data processing and analysis. Computational resources were supplied by the project "e-Infrastruktura CZ" (e-INFRA CZ LM2018140 ) supported by the Ministry of Education, Youth and Sports of the Czech Republic.

%
%
%

\bibliographystyle{tsd1108a}
\bibliography{tsd1108a}

\end{document}